\documentclass[10pt,letterpaper]{article}

\usepackage{cogsci}

\cogscifinalcopy 

\usepackage{cmap}
\usepackage[T1]{fontenc}
\usepackage[american]{babel}
\usepackage{csquotes}
\usepackage{newtxtext,newtxmath}  

\usepackage[
  backend=biber,
  style=apa,
  natbib=true,
  annotation=false,
]{biblatex}
\usepackage{float} 

\usepackage[hidelinks]{hyperref}
\usepackage{graphicx}
\usepackage{booktabs}
\usepackage{array}
\newcolumntype{C}{>{\centering\arraybackslash}p{0.065\textwidth}}

\title{Learning to Trade Like an Expert: Cognitive Fine-Tuning for Stable Financial Reasoning in Language Models}

\author[1]{\mbox{Yuchen Pan (ypan@link.cuhk.edu.hk)}}
\author[1]{\mbox{Soung Chang Liew (soung@ie.cuhk.edu.hk)}}
\affil[1]{Department of Information Engineering, The Chinese University of Hong Kong}

\begin{document}

\maketitle

\begin{abstract}
Recent deployments of large language models (LLMs) as autonomous trading agents raise questions about whether financial decision-making competence generalizes beyond specific market patterns and how it should be trained and evaluated in noisy markets lacking ground truth. We propose a structured framework for training and evaluating such models. Central to our approach is a curated, multiple-choice question (MCQ) dataset derived from classic textbooks and historical markets, verified by an AI committee, enriched with structured reasoning traces, and augmented to reduce shortcut learning. To evaluate whether performance on isolated MCQs generalizes to real-world trading, we introduce a two-stage protocol combining test-set evaluation with an MCQ-based chronological trading simulation. Extensive evaluations across market regimes provide statistically robust evidence that open models trained with our framework exhibit competitive, risk-aware behavior over time, outperform open-source baselines, and approach frontier-model performance at smaller scale. We release the dataset and evaluation framework to support further research.

\end{abstract}

\section{Introduction} 
\label{Section_I}
In late October 2025, financial artificial intelligence reached a new milestone with the launch of the Alpha Arena initiative.\footnote{https://nof1.ai/} For the first time, frontier LLMs were entrusted with real capital to trade autonomously in live U.S. equity and cryptocurrency markets, competing directly against one another.

Despite the widespread attention it received, the initiative raises fundamental questions about autonomous decision-making in finance. \textbf{Question 1}: Does the observed profitability reflect genuine domain competence that generalizes across diverse market conditions, or is it merely an artifact of specific market patterns tied to particular time periods? \textbf{Question 2}: If such competence exists, can it be achieved more reliably through specialized trading models rather than general-purpose LLMs? Even if these agents demonstrate generalizable trading competence, deploying them in institutional finance presents additional challenges related to training, governance, and control. \textbf{Challenge 1}: Reliance on externally hosted, closed-source models introduces significant risks, including data privacy concerns, regulatory compliance issues, and the potential leakage of proprietary trading strategies—factors that increasingly motivate a shift toward locally deployed models. \textbf{Challenge 2}: Financial markets are inherently noisy and lack clear "correct" answers, making supervised learning on raw price data prone to overfitting spurious correlations rather than capturing causal decision principles. As a result, financial institutions have a compelling need for alternative training methods and deployment platforms that provide structured, high-quality guidance while fostering reasoning and judgment under uncertainty.

Existing literature offers few direct precedents to address these questions. Most open-source financial language models are not designed for autonomous execution: Discriminative models \citep{araci2019finbert, huang2023finbert, liu2021finbert} primarily focus on sentiment analysis, while generative models \citep{wu2023bloomberggpt, yang2023fingpt, zhang2023instructfingpt} are largely aimed at semantic retrieval or advisory tasks \citep{yang2023investlm, chen2023disc}. Recent execution-oriented frameworks \citep{zhou2024finrobot, zhang2024ai, xiao2025tradingagentsmultiagentsllmfinancial} often rely on complex multi-agent systems, which obscure the model’s intrinsic reasoning process. Furthermore, existing financial datasets and benchmarks \citep{chen2021finqa, li2024alphafin} predominantly emphasize data extraction or price prediction rather than decision-making logic. To our knowledge, no publicly available resource explicitly provides a framework for financial decision logic.

\begin{figure*}[t]
  \begin{center}
    \includegraphics[width=\textwidth]{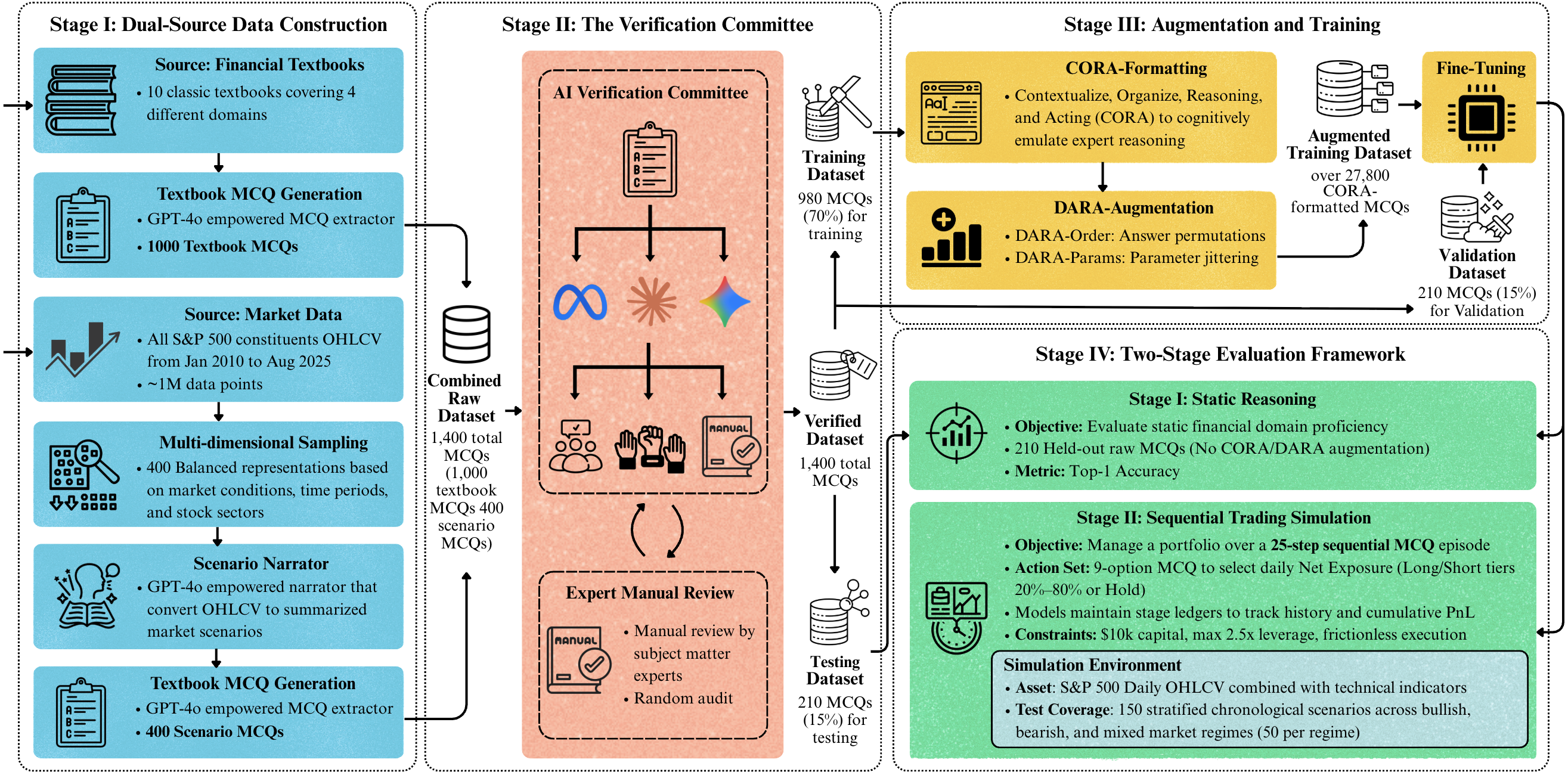}
  \end{center}
  \caption{Overview of the proposed framework.}
  \label{fig:workflow}
\end{figure*}

Building on these questions and challenges, we make three contributions toward training and evaluating robust, locally deployable financial agents. \textbf{First}, in response to Question 2 and Challenge 2, we introduce a curated MCQ dataset for training and evaluating autonomous trading agents. The dataset is derived from classic textbooks and historical market scenarios grouped by market regime, time period, and sector. Each item is verified by an AI committee composed of three LLMs for factual and logical errors, annotated with a four-step chain-of-thought (CoT) trace, and augmented by shuffling answer options and varying technical indicator parameters to reduce shortcut learning. \textbf{Second}, to investigate Question 1, we propose a two-stage evaluation protocol: The first stage measures performance on isolated MCQs in a held-out test set, while the second stage evaluates whether this competence generalizes to better real-world trading behavior through a chronological, MCQ-based trading simulation under realistic portfolio constraints. Extensive evaluations across bullish, bearish, and mixed market regimes provide statistically robust evidence that models trained with our framework translate strong first-stage performance into consistent, risk-aware real-world trading behavior, enabling smaller-scale open models to perform competitively with controlled downside risk. \textbf{Finally}, to bridge the deployment gap in Challenge 1, we open source the full training and evaluation pipeline at \url{https://github.com/anonymous/DoubleBlind} to support customization and further research on locally deployed financial agents. Figure~\ref{fig:workflow} summarizes the proposed framework.

\section{Related Works} 
\label{Section_II}
\noindent\hspace*{\parindent}\textbf{Financial LLM Evolution}\hspace{0.8em} Financial language models have progressed from task-specific classifiers to more general systems used in financial analysis and trading applications \citep{Li2023Large, Ding2024Large, Dong2025Large}. Early models such as FinBERT \citep{araci2019finbert, liu2021finbert, huang2023finbert} were primarily designed for sentiment analysis, mapping financial news and social media to trading signals. For broader financial applications beyond sentiment analysis, closed-source models like BloombergGPT \citep{wu2023bloomberggpt} were pre-trained for tasks such as financial question answering and investment advisory. Open-source models including FinGPT \citep{yang2023fingpt, zhang2023instructfingpt}, InvestLM \citep{yang2023investlm}, and DISC-FinLLM \citep{chen2023disc} have further adapted open models to these tasks through parameter-efficient post-training. More recently, LLMs have been integrated directly into trading pipelines via hierarchical planning \citep{zhou2024finrobot} and multi-agent designs \citep{zhang2024ai, yu2024fincon, xiao2025tradingagentsmultiagentsllmfinancial}. In this work, we combine open-weight post-training with direct trading execution, relying on a single model instead of complex planning or multi-agent architectures.

\textbf{Structured Reasoning Strategies}\hspace{0.8em} Another line of work focuses on improving the reliability of LLM outputs, which is especially important in finance. Prior studies showed that standard LLMs can produce hallucinated explanations that are not causally aligned with their final answers \citep{webb2023emergent, binz2023using, turpin2023language}. To mitigate this, inference-time methods prompt models to produce explicit intermediate steps, including linear reasoning chains  \citep{wei2022chain, wang2022self}, branching sets of reasoning paths \citep{yao2023tree, besta2024graph}, and iterative self-critique and revision \citep{shinn2023reflexion, madaan2023self}. Training-time methods encourage similar behavior by fine-tuning models on annotated reasoning traces \citep{hsieh2023distilling, hao2023reasoning, lightman2023let, uesato2022solving} or by sampling multiple candidate outputs and fine-tuning the model on the highest-quality ones \citep{zelikman2022star, zelikman2024quiet}. Our work follows the training-time paradigm by fine-tuning models on annotated reasoning traces.

\textbf{Data Curation and Benchmarks}\hspace{0.8em} Training and evaluating financial models is challenging due to market noise and the lack of definitive ground truth. Prior work showed that problems with clear structure and explicit intermediate steps can yield reliable signals for learning and assessment \citep{gunasekar2023textbooks, 10993716}. This principle appears in financial datasets like FinQA and ConvFinQA \citep{chen2021finqa, chen2022convfinqa}, which emphasize financial text and numerical tables, as well as MultiHiertt and BizBench \citep{zhao2022multihiertt, krumdick2024bizbench}, which focus on longer business documents. In parallel, datasets based on real market data—including AlphaFin \citep{li2024alphafin}, FinBen \citep{xie2024finben}, and StockBench \citep{Chen2025StockBench}—emphasize time-series signals and typically frame tasks as information retrieval or market prediction. Our work adopts both directions by deriving problems from real market data and annotating them with explicit reasoning steps.

\section{Cognitive Financial Reasoning Dataset} 
\label{Section_III}
\subsection{Dual-Source Data Construction}
To support structured training and evaluation of trading agents, we constructed the Cognitive Financial Reasoning Dataset. We used an MCQ format to define a clear decision space that supports controlled analysis of model choices. As illustrated in Figure~\ref{fig:workflow} (Stage~1), our data construction pipeline is designed to mimic the \emph{learning process of human experts}: first acquiring foundational principles from classic \textbf{Textbooks}, and then refining judgment through exposure to dynamic \textbf{Market Data}. The full list of source textbooks, the prompt templates used to generate MCQs from textbooks and market data, and the final datasets are available in our open-source repository.

\subsubsection{Textbook Knowledge Extraction}
To build a reliable foundation for financial reasoning, we first curated ten classic textbooks spanning four key domains essential to professional trading practice: Technical Analysis, Quantitative Trading, Macroeconomics, and Trading Psychology. We used GPT-4o \citep{hurst2024gpt} to identify core concepts within each textbook section and then converted them into discrete MCQs that emphasize reasoning about trading decisions rather than simple recall of definitions or facts. For example, for a chapter introducing indicator divergence, we generated an MCQ describing a price downtrend with a bullish RSI divergence and asked for the most prudent trading action. We generated exactly 100 MCQs per textbook, yielding a balanced theoretical core of 1,000 textbook-derived questions. 

\subsubsection{Market Scenario Extraction}
To help models make decisions in noisy, real-world markets, we created 400 additional scenario-based MCQs using roughly 15 years of S\&P 500 historical data (2010--2025). We started with about one million data records and split the time series into fixed-length windows. Each window was annotated with three tag categories (listed below), and we sampled roughly evenly across the resulting tag tuples to ensure balanced coverage across:
\begin{enumerate}
    \item Market regimes (e.g., strong uptrends, high volatility)
    \item Time periods (e.g., post-crisis recovery)
    \item Market sectors (e.g., healthcare, energy)
\end{enumerate}

We used a two-step GPT-4o pipeline to convert numeric signals into text. For each price window, the \textbf{Narrator} summarized the data and 38 technical indicators into a natural-language description (e.g., \textit{``TSLA shows overbought conditions with \%K=97.6; a slightly negative MACD=-0.12 suggests waning bullish momentum\ldots''}). The \textbf{Generator} then turned the summary into an MCQ asking for the best trading action (e.g., enter now). Together with the textbook questions, this produced a pre-verification dataset of 1{,}400 MCQs.

\subsection{The Verification Committee}
To mitigate hallucinations and over-reliance on the single generator model GPT-4o, we introduce a verification stage (Figure~\ref{fig:workflow}, Stage~2) using an \textbf{AI committee} of three prior-generation frontier LLMs: Gemini-2.5-Flash \citep{comanici2025gemini}, Claude-3.7-Sonnet \citep{anthropic2025claude37}, and Llama-4-Scout \citep{meta2025llama}. This setup is similar to using multiple human judges to reduce individual errors. We use earlier-generation models because our later experiments evaluate newer frontier models as baselines; keeping the committee separate helps avoid overlap between verification and evaluation. To safeguard label quality, expert human review is used as the final source of labels for difficult cases.

Each of the 1,400 raw MCQs was independently evaluated by all AI committee members under a standardized prompt (released in our repository). For every item, each model was required to choose an answer, provide a brief rationale, and flag any suspected factual errors; when the information was insufficient, it was instructed to output \texttt{UNCERTAIN} rather than guess. Final labels were assigned via a strict consensus rule:

\begin{itemize}
\item \textbf{Unanimous agreement (3/3):} 1{,}139 items (81.4\%) were retained as high-confidence labels.
\item \textbf{Majority agreement (2/3):} 158 items (11.3\%) were accepted, with the committee’s majority answer replacing the original generator’s label.
\item \textbf{No agreement:} The remaining 103 items were reviewed by domain experts; 96 reached consensus after revising the MCQs to clarify wording and correct factual issues, and 7 were further simplified for clarity.
\end{itemize}

In a follow-up random audit of 30 verified items, human experts agreed with the committee’s labels on 29 of them (96.67\%), suggesting that the AI committee can approximate expert verification at scale.

\subsection{CORA: Cognitive Chain-of-Thought}
For humans, expertise in financial decision-making is not acquired by memorizing isolated answers, but through a progressive learning process of first reading market context, then framing the decision, weighing alternatives, and finally translating judgment into a concrete trading plan. To mirror this learning trajectory, we introduce \textbf{CORA (Contextualize–Organize–Reason–Act)}, a four-stage CoT template that expands each verified MCQ into:

\begin{itemize}
\item \textbf{Contextualize}: Synthesizes the market state into a coherent representation, integrating trends, volatility, technical indicators, and relevant background factors.
\item \textbf{Organize}: Defines the decision frame by specifying the trading objective and key constraints, reflecting how experts simplify complex environments.
\item \textbf{Reason}: Justifies the chosen option and explicitly explains why the other options are less appropriate, encouraging contrastive reasoning.
\item \textbf{Act}: Translates the conclusion into a concrete trading plan, including direction, entry conditions, and risk controls.
\end{itemize}

\begin{figure}[t]
    \centering
    \includegraphics[width=1\linewidth]{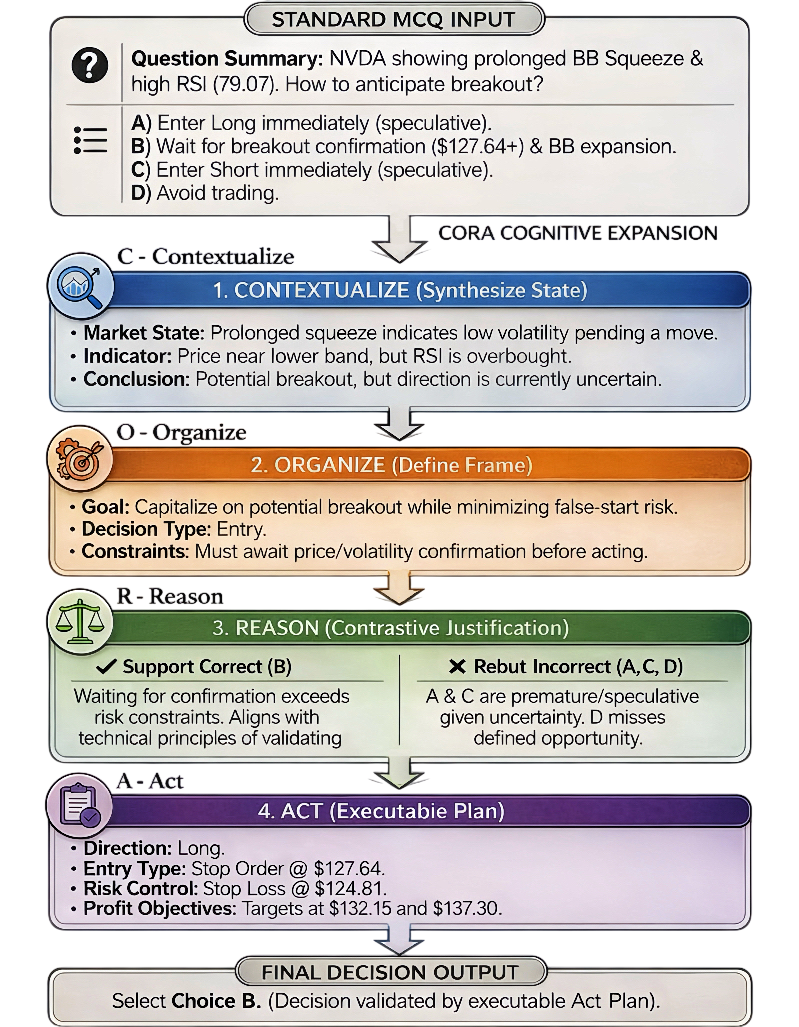}
    \caption{A simplified example of CORA cognitive expansion.}
    \label{fig:cora_example}
\end{figure}

Figure~\ref{fig:cora_example} illustrates a simplified example of how a single MCQ is expanded into CORA format. We use GPT-4o with standardized prompts to generate the four-stage traces. Manual audits of 50 randomly selected items confirm that the generated reasoning is clear and consistent with the underlying question. By turning simple answer labels into structured reasoning traces, CORA shifts the learning signal from \textit{which} decision is correct to \textit{how} decisions are formed under uncertainty. Instead of treating financial markets as if they had definitive ground truth, CORA encodes expert-like patterns—such as attention to volatility and explicit risk management—that resemble how human traders reason in practice.

\subsection{DARA: Dual-Axis Robustness Augmentation}
Analysis of the initial dataset revealed two systematic biases introduced during generation. First, correct answers were not evenly spread across options: In the 1,400 MCQs, options A and B covered 76.9\% of correct labels, while option D appeared in only 4.2\% of cases. Second, many textbook-derived questions reused the same indicator settings (e.g., repeatedly using a 14-day moving average). These patterns can lead to shortcut learning, where models exploit superficial cues such as \textit{“A is usually correct”} or fixed parameter values instead of reasoning about the underlying scenario.

To address these issues, we introduce \textbf{Dual-Axis Robustness Augmentation (DARA)}, which preserves the CORA structure while reducing reliance on answer position and specific numeric settings. It operates along two dimensions:

\begin{itemize}
\item \textbf{DARA-Order}: Addresses positional bias by generating all $4! = 24$ permutations of the answer choices for each MCQ.
\item \textbf{DARA-Params}: Reduces overfitting to specific indicator values by introducing controlled variations within common practical ranges (e.g., varying RSI-14 to RSI-21).
\end{itemize}

When applied to the 980 training MCQs, DARA-Order yields a $24\times$ expansion, while DARA-Params introduces an additional $\sim$1.18$\times$ increase, resulting in approximately 27,800 training examples. We do not apply CORA or DARA to the validation or test sets; they remain unchanged to provide a consistent benchmark for measuring generalization.

Together, CORA and DARA form our reasoning-focused augmentation strategy: CORA makes the decision logic explicit, while DARA discourages reliance on answer position and fixed numeric settings. The prompt templates for both stages are included in our open-source repository.

\section{Two-Stage Evaluation Framework} 
\label{Section_IV}
\begin{table*}[t]
\begin{center}
\caption{Stage I and Stage II performance. Stage I reports test accuracy. Stage II reports overall average return (Ret, \%), mean net exposure (Exp; average absolute fraction of capital allocated), and turnover (Tur; cumulative absolute change in net exposure per episode, \%), together with regime-specific average returns and downside risk measured by CVaR-5\% (\%), defined as the mean return of the worst-performing 5\% episodes. Stage II results are not reported for \emph{w/o CORA} due to degenerate behavior.}
\label{tab:hero_results}
\vskip 0.12in
\small

\setlength{\aboverulesep}{0.365ex}
\setlength{\belowrulesep}{0.365ex}

\begin{tabular*}{\textwidth}{
@{\extracolsep{\fill}}
l c ccc cc cc cc
}
\toprule
& \multicolumn{1}{c}{\textbf{Stage I}}
& \multicolumn{3}{c}{\textbf{Stage II: Overall}}
& \multicolumn{6}{c}{\textbf{Stage II: Regime-Specific}} \\
\cmidrule(lr){2-2}
\cmidrule(lr){3-5}
\cmidrule(lr){6-11}

& \textbf{Static}
& \multicolumn{3}{c}{\textbf{(150 Episodes)}}
& \multicolumn{2}{c}{\textbf{Bullish}}
& \multicolumn{2}{c}{\textbf{Bearish}}
& \multicolumn{2}{c}{\textbf{Mixed}} \\
\cmidrule(lr){6-7}
\cmidrule(lr){8-9}
\cmidrule(lr){10-11}

\textbf{Model / Method}
& \textbf{Acc}
& \textbf{Ret}
& \textbf{Exp}
& \textbf{Tur}
& \textbf{Ret}
& \textbf{CVaR}
& \textbf{Ret}
& \textbf{CVaR}
& \textbf{Ret}
& \textbf{CVaR} \\
\midrule

\multicolumn{11}{@{}l}{\textit{\textbf{Frontier Models}}} \\
GPT-5.1 \citep{OpenAI_2025} & 80.95 & 7.08 & 0.68 & 157.7 & 10.52 & -13.1 & 9.81 & -3.41 & 0.91 & -7.89 \\
Claude-4.5-Sonnet \citep{Claude_2025} & 86.19 & \textbf{12.04} & 0.82 & 54.5 & 21.26 & 3.80 & \textbf{14.59} & \textbf{-0.24} & 0.26 & -9.60 \\
DeepSeek-v3.1 \citep{DeepSeek2024DeepSeek} & \textbf{87.14} & 10.56 & 1.30 & 64.8 & \textbf{45.84} & 3.75 & -14.45 & -33.43 & 0.29 & -4.23 \\
\midrule

\multicolumn{11}{@{}l}{\textit{\textbf{Open-Source Baselines}}} \\
Llama-3.1-8B-Instruct & 57.62 & 6.23 & 1.11 & 284.6 & 24.61 & -14.35 & -4.93 & -32.60 & -0.99 & -14.68 \\
Qwen3-8B \citep{yang2025qwen3} & 71.90 & 6.33 & 0.87 & 185.2 & 10.71 & -21.42 & 8.05 & -25.23 & 0.24 & -13.96 \\
Mistral-7B-Instruct-v0.3 \citep{Jiang2023Mistral} & 64.29 & 0.43 & 1.19 & 315.2 & 20.12 & -11.17 & -19.41 & -47.57 & 0.57 & -12.99 \\
\midrule

\multicolumn{11}{@{}l}{\textit{\textbf{Ablation Analysis}}} \\
Ours w/o CORA, DARA \& Verification & 70.95 & 1.36 & 1.22 & 141.2 & 27.57 & 1.04 & -14.25 & -28.36 & 0.19 & -5.42 \\
Ours w/o DARA & 76.67 & 3.07 & 0.96 & \textbf{29.5} & 22.39 & 3.81 & -13.39 & -29.37 & 0.22 & \textbf{-1.69} \\
Ours w/o CORA & 46.19 & — & — & — & — & — & — & — & — & — \\
\midrule

\textbf{Ours (Full Model)} & 82.38 & 7.64 & \textbf{1.70} & 82.7 & 43.66 & \textbf{8.19} & -21.67 & -36.94 & \textbf{0.93} & -2.64 \\
\bottomrule
\end{tabular*}
\end{center}
\end{table*}

We propose a two-stage evaluation framework for financial reasoning that evaluates models on both \textbf{static reasoning}, where each decision is an independent MCQ response, and \textbf{sequential trading behavior}, where decisions accumulate over time in a live portfolio. At test time, models receive only the MCQ prompt (market scenario and answer choices in Stage I, plus the current portfolio ledger state in Stage II), and no gold CORA traces are provided. Models are prompted to generate their own reasoning (CORA-style for CORA-trained variants; standard CoT for others) before giving a final answer.

\subsection{Stage I: Static Reasoning}
Stage I follows a standard supervised evaluation paradigm. We evaluate each model on a test set of 210 MCQs. Each question is answered independently to assess how well the model performs on isolated financial decisions. 

\subsection{Stage II: Sequential Trading Simulation}
While Stage I evaluates isolated MCQ answering accuracy, static performance alone is insufficient for assessing practical trading competence. In real financial markets, decisions unfold sequentially, affecting future capital, exposure, and risk. Models that perform well on static tasks may nevertheless behave unstably or inconsistently when their decisions are chained together in a portfolio.

To address this, Stage II introduces a \textbf{chronological evaluation protocol} that preserves the MCQ format of Stage I while placing decisions within a continuous trading simulation. In Stage~II, the model is asked to manage a portfolio over a fixed 25-step episode (approximately one trading month). At every step, it receives the current market scenario along with a state ledger documenting prior actions, current positions, and cumulative profit and loss. This running context allows each decision to condition on earlier outcomes. The model then answers an MCQ by selecting one of nine target net exposures for the next session: long or short at 20\%, 40\%, 60\%, or 80\% of allowable exposure, or a neutral hold.

Stage II also differs from open-ended agent setups (such as the Alpha Arena initiative described in the Introduction), where models operate in large, unstructured action spaces using free-form text, making it hard to separate reasoning quality from parsing errors or prompt-specific artifacts. In contrast, Stage II restricts the action space to a small set of interpretable exposure choices, making it easier to relate each decision to a clear change in the portfolio.

Episodes are built from daily S\&P 500 data and standard technical indicators using the same pipeline as the scenario-based MCQs, except that the answer choices are always fixed to the same nine exposure options mentioned above. Models receive no access to future information. We enforce an initial capital of \$10{,}000, a maximum gross leverage of 2.5$\times$, and frictionless execution to isolate decision quality from trading frictions such as bid–ask spreads and slippage.

To enable \textbf{statistically robust evaluation} beyond specific market patterns, we run 150 non-overlapping episodes across bullish, bearish, and mixed market regimes (50 per regime), for a total of 3,750 MCQ steps. Regimes are defined by buy-and-hold returns over each 25-day window, and episodes are selected using a round-robin strategy over time to avoid overrepresenting any particular period.

\section{Experiments and Results} 
\label{Section_V}
\subsection{Experimental Setup}
We fine-tune Llama-3.1-8B-Instruct \citep{Meta_2024b} on our Cognitive Financial Reasoning Dataset using Q-LoRA \citep{dettmers2023qlora}. The model is selected for its open-weight availability and strong instruction-following performance at the 8B scale, while Q-LoRA enables efficient fine-tuning under resource-constrained settings. Training uses the standard Llama-3.1 chat template: "system messages" in the template define the task, "user messages" provide market scenarios and MCQ options, and "assistant messages" output a CORA-structured reasoning trace and the final selected option (i.e., the target completion). Each MCQ is serialized as a single chat turn, and cross-entropy is applied only to assistant tokens (system/user tokens are masked), encouraging learned reasoning and answer selection rather than prompt reproduction.

We benchmark against frontier LLMs and size-matched open-source models. Frontier LLMs (e.g., GPT-5.1 \citep{OpenAI_2025}), similar to those used in the Alpha Arena initiative, are evaluated via APIs and serve as reference points for high-capacity models on financial tasks. Open-source models at comparable scales (e.g., Qwen3-8B \citep{yang2025qwen3}) serve as size-matched baselines for locally deployable systems. For both Stage I and Stage II (including all 150 trading episodes), we set the decoding temperature to 0 for all baseline and locally fine-tuned models to ensure deterministic and directly comparable outputs. To isolate the effects of the components in our method, we run ablations that remove verification, CORA, and DARA. For a fair comparison, CORA-trained models are prompted to produce the full four-stage CORA output before giving a final answer, whereas other models are prompted to generate standard CoT with a similar output length. This setup ensures that performance differences primarily reflect the learned reasoning structure, rather than variations in the amount of intermediate text generated.

\subsection{Results and Discussion}
Table~\ref{tab:hero_results} reports performance for both Stage~I and Stage~II.

\textbf{Stage~I}\hspace{0.8em} Frontier models achieve high MCQ accuracy (80.95–87.14\%), while open-source baselines lag behind. Our full model attains 82.38\% accuracy, outperforming all open-source baselines under the same conditions and approaching the frontier range despite its smaller size and lack of external tools. Ablation results clarify the sources of these gains. Standard supervised fine-tuning on raw MCQs (\textit{without CORA, DARA and verification}) achieves 70.95\% accuracy, whereas the full pipeline improves this by over 11 percentage points. Removing DARA lowers accuracy to 76.67\% and increases shortcut behavior. Removing CORA has a much stronger effect: Accuracy drops to 46.19\%, and the model overfits to the DARA permutations, collapsing into repeatedly selecting the same option. These results suggest that our structured training is crucial when learning from noisy financial data.

\textbf{Stage~II}\hspace{0.8em} We next test whether single-step performance carries over to sequential trading using metrics that capture profitability, capital use, behavioral stability, and downside risk (defined in the table caption). In the overall evaluation (150 episodes), our full model achieves an average return of 7.64\%, outperforming all open-source baselines and approaching frontier-level performance. It also maintains the highest mean net exposure with moderate turnover, indicating confident positioning without excessive trading. Several baselines either trade too frequently or keep exposure too low, suggesting difficulty in maintaining stable policies. Results for the \textit{w/o CORA} variant are omitted because it exhibits degenerate behavior, selecting identical exposure levels across all episodes (consistent with the Stage~I findings).

\begin{figure}[t]
    \centering
    \includegraphics[width=1\linewidth]{Figures/wealth_lines_bullish.png}
    \caption{\textbf{Bullish-regime wealth trajectories.} Solid lines show mean portfolio value over 50 simulation runs for Claude-4.5-Sonnet, Qwen3-8B, and our Full Model; shaded regions show \(\pm 1\) standard deviation.}
    \label{fig:wealth_bullish}
\end{figure}

\textbf{Regime Sensitivity}\hspace{0.8em} Breaking the results down by market regime shows how model behavior changes across conditions. In bullish regimes, our model delivers strong returns (43.66\%) and achieves the best 5\% CVaR, indicating controlled downside risk. In mixed regimes, it again attains the highest average return with limited tail risk, suggesting it can handle uncertainty without taking excessive downside risk. In bearish regimes, however, it underperforms most baselines in both average return and tail risk. Overall, these results suggest an assertive long-biased policy—also seen in DeepSeek-v3.1—that works well in favorable or mixed conditions but is not well adapted to sustained bearish markets. Figure~\ref{fig:wealth_bullish} visualizes portfolio value trajectories in bullish-regime episodes. For clarity, we plot three representative models with the highest overall return in each category. Shaded regions indicate dispersion across 50 runs within ±1 standard deviation. The Full Model exhibits a consistently higher distribution of outcomes across the trading horizon, suggesting that the observed advantage reflects a systematic behavioral improvement rather than a purely random fluctuation.

\textbf{Stage~I vs.\ Stage~II Relationship}\hspace{0.8em} Stage I accuracy and Stage II trading performance are related but not equivalent. Among frontier models, the highest Stage I accuracy comes from DeepSeek-v3.1 (87.14\% accuracy, 10.56\% return), but the best average return is achieved by Claude-4.5-Sonnet (86.19\%, 12.04\%). This shows that stronger single-step accuracy does not always translate into better portfolio outcomes. However, within our variants, improvements in Stage I accuracy do track improvements in Stage II performance. Moving from \textit{standard fine-tuning} (70.95\%, 1.36\%) to \textit{without DARA} (76.67\%, 3.07\%), and then to the \textit{full model} (82.38\%, 7.64\%), increases both test accuracy and average return. The ablations also highlight that removing DARA keeps Stage I accuracy relatively high but reduces trading performance. Overall, high static accuracy appears necessary but not sufficient for strong trading behavior, and the CORA- and DARA-based training pipeline helps translate single-step reasoning into more consistent portfolio decisions over time.

\section{Conclusion} 
\label{Section_VI}
This paper investigates whether LLMs can acquire generalizable financial decision-making competence beyond specific market patterns. We introduce the Cognitive Financial Reasoning Dataset, along with an AI verification committee, CORA, a four-stage reasoning template, and DARA, a data augmentation method that reduces shortcut learning in noisy market settings. To evaluate whether single-step reasoning transfers to real-world trading, we propose a two-stage evaluation framework that combines static MCQ testing with a chronological trading simulation under portfolio constraints.

Our results show that models trained with this pipeline outperform open-source baselines and approach frontier-model performance at a smaller scale in bullish and mixed regimes, but require further improvement in bearish regimes.

Future work includes strengthening robustness in bearish markets, extending the framework to multi-asset settings, adding signals such as market sentiment and trading friction, and studying how explicit reasoning traces can support interpretability, governance, and safe real-world deployment.

\section{Acknowledgment}
The authors used large language models to assist with code development and with drafting and editing the manuscript.

\printbibliography

@inproceedings{Li2023Large,
	author = {Li, Yinheng and Wang, Shaofei and Ding, Han and Chen, Hang},
	booktitle = {4th {ACM} {International} {Conference} on {AI} in {Finance}},
	doi = {10.1145/3604237.3626869},
	year = {2023},
	month = {nov 25},
	pages = {374--382},
	organization = {ACM},
	title = {Large {Language} {Models} in {Finance}: A {Survey}},
	url = {http://dx.doi.org/10.1145/3604237.3626869},
}

@article{Ding2024Large,
	author = {Ding, Han and Li, Yinheng and Wang, Junhao and Chen, Hang},
	doi = {10.48550/ARXIV.2408.06361},
	year = {2024},
	publisher = {arXiv},
	title = {Large {Language} {Model} {Agent} in {Financial} {Trading}: A {Survey}},
	url = {https://arxiv.org/abs/2408.06361},
}

@inproceedings{Dong2025Large,
	author = {Dong, Yifei and Wu, Fengyi and Zhang, Kunlin and Dai, Yilong and Zhang, Sanjian and Ye, Wanghao and Chen, Sihan and Cheng, Zhi-Qi},
	booktitle = {Findings of the {Association} for {Computational} {Linguistics}: EMNLP 2025},
	doi = {10.18653/v1/2025.findings-emnlp.972},
	year = {2025},
	pages = {17889--17907},
	organization = {Association for Computational Linguistics},
	title = {Large {Language} {Model} {Agents} in {Finance}: A {Survey} {Bridging} {Research}, {Practice}, and {Real}-{World} {Deployment}},
	url = {http://dx.doi.org/10.18653/v1/2025.findings-emnlp.972},
}

@article{araci2019finbert,
  title={Finbert: Financial sentiment analysis with pre-trained language models},
  author={Araci, Dogu},
  journal={arXiv preprint arXiv:1908.10063},
  year={2019}
}

@inproceedings{liu2021finbert,
  title={Finbert: A pre-trained financial language representation model for financial text mining},
  author={Liu, Zhuang and Huang, Degen and Huang, Kaiyu and Li, Zhuang and Zhao, Jun},
  booktitle={Proceedings of the twenty-ninth international conference on international joint conferences on artificial intelligence},
  pages={4513--4519},
  year={2021}
}

@article{huang2023finbert,
  title={FinBERT: A large language model for extracting information from financial text},
  author={Huang, Allen H and Wang, Hui and Yang, Yi},
  journal={Contemporary Accounting Research},
  volume={40},
  number={2},
  pages={806--841},
  year={2023},
  publisher={Wiley Online Library}
}

@article{wu2023bloomberggpt,
  title={Bloomberggpt: A large language model for finance},
  author={Wu, Shijie and Irsoy, Ozan and Lu, Steven and Dabravolski, Vadim and Dredze, Mark and Gehrmann, Sebastian and Kambadur, Prabhanjan and Rosenberg, David and Mann, Gideon},
  journal={arXiv preprint arXiv:2303.17564},
  year={2023}
}

@article{wei2022chain,
  title={Chain-of-thought prompting elicits reasoning in large language models},
  author={Wei, Jason and Wang, Xuezhi and Schuurmans, Dale and Bosma, Maarten and Xia, Fei and Chi, Ed and Le, Quoc V and Zhou, Denny and others},
  journal={Advances in neural information processing systems},
  volume={35},
  pages={24824--24837},
  year={2022}
}

@article{yang2023fingpt,
  title={FinGPT: Open-Source Financial Large Language Models},
  author={Yang, Hongyang and Liu, Xiao-Yang and Wang, Christina Dan},
  journal={FinLLM Symposium at IJCAI 2023},
  year={2023}
}

@article{zhang2023instructfingpt,
      title={Instruct-FinGPT: Financial Sentiment Analysis by Instruction Tuning of General-Purpose Large Language Models}, 
      author={Boyu Zhang and Hongyang Yang and Xiao-Yang Liu},
      journal={FinLLM Symposium at IJCAI 2023},
      year={2023}
}

@misc{yang2023investlm,
      title={InvestLM: A Large Language Model for Investment using Financial Domain Instruction Tuning}, 
      author={Yi Yang and Yixuan Tang and Kar Yan Tam},
      year={2023},
      eprint={2309.13064},
      archivePrefix={arXiv},
      primaryClass={q-fin.GN}
}

@article{chen2023disc,
  title={DISC-FinLLM: A Chinese Financial Large Language Model based on Multiple Experts Fine-tuning},
  author={Chen, Wei and Wang, Qiushi and Long, Zefei and Zhang, Xianyin and Lu, Zhongtian and Li, Bingxuan and Wang, Siyuan and Xu, Jiarong and Bai, Xiang and Huang, Xuanjing and Wei, Zhongyu},
  journal={arXiv preprint arXiv:2310.15205},
  year={2023}
}

@inproceedings{
zhou2024finrobot,
title={FinRobot: {AI} Agent for Equity Research and Valuation with Large Language Models},
author={Tianyu Zhou and Pinqiao Wang and Yilin Wu and Hongyang Yang},
booktitle={ICAIF 2024: The 1st Workshop on Large Language Models and Generative AI for Finance},
year={2024}
}

@article{zhang2024ai,
  title={When AI Meets Finance (StockAgent): Large Language Model-based Stock Trading in Simulated Real-world Environments},
  author={Zhang, Chong and Liu, Xinyi and Jin, Mingyu and Zhang, Zhongmou and Li, Lingyao and Wang, Zhengting and Hua, Wenyue and Shu, Dong and Zhu, Suiyuan and Jin, Xiaobo and others},
  journal={arXiv preprint arXiv:2407.18957},
  year={2024}
}

@misc{xiao2025tradingagentsmultiagentsllmfinancial,
      title={TradingAgents: Multi-Agents LLM Financial Trading Framework}, 
      author={Yijia Xiao and Edward Sun and Di Luo and Wei Wang},
      year={2025},
      eprint={2412.20138},
      archivePrefix={arXiv},
      primaryClass={q-fin.TR},
      url={https://arxiv.org/abs/2412.20138}, 
}

@article{yu2024fincon,
  title={Fincon: A synthesized llm multi-agent system with conceptual verbal reinforcement for enhanced financial decision making},
  author={Yu, Yangyang and Yao, Zhiyuan and Li, Haohang and Deng, Zhiyang and Jiang, Yuechen and Cao, Yupeng and Chen, Zhi and Suchow, Jordan and Cui, Zhenyu and Liu, Rong and others},
  journal={Advances in Neural Information Processing Systems},
  volume={37},
  pages={137010--137045},
  year={2024}
}

@article{yao2023tree,
  title={Tree of thoughts: Deliberate problem solving with large language models},
  author={Yao, Shunyu and Yu, Dian and Zhao, Jeffrey and Shafran, Izhak and Griffiths, Tom and Cao, Yuan and Narasimhan, Karthik},
  journal={Advances in neural information processing systems},
  volume={36},
  pages={11809--11822},
  year={2023}
}

@article{webb2023emergent,
  title={Emergent analogical reasoning in large language models},
  author={Webb, Taylor and Holyoak, Keith J and Lu, Hongjing},
  journal={Nature Human Behaviour},
  volume={7},
  number={9},
  pages={1526--1541},
  year={2023},
  publisher={Nature Publishing Group UK London}
}

@article{binz2023using,
  title={Using cognitive psychology to understand GPT-3},
  author={Binz, Marcel and Schulz, Eric},
  journal={Proceedings of the National Academy of Sciences},
  volume={120},
  number={6},
  pages={e2218523120},
  year={2023},
  publisher={National Academy of Sciences}
}

@article{shinn2023reflexion,
  title={Reflexion: Language agents with verbal reinforcement learning},
  author={Shinn, Noah and Cassano, Federico and Gopinath, Ashwin and Narasimhan, Karthik and Yao, Shunyu},
  journal={Advances in Neural Information Processing Systems},
  volume={36},
  pages={8634--8652},
  year={2023}
}

@article{madaan2023self,
  title={Self-refine: Iterative refinement with self-feedback},
  author={Madaan, Aman and Tandon, Niket and Gupta, Prakhar and Hallinan, Skyler and Gao, Luyu and Wiegreffe, Sarah and Alon, Uri and Dziri, Nouha and Prabhumoye, Shrimai and Yang, Yiming and others},
  journal={Advances in Neural Information Processing Systems},
  volume={36},
  pages={46534--46594},
  year={2023}
}

@inproceedings{lightman2023let,
  title={Let's verify step by step},
  author={Lightman, Hunter and Kosaraju, Vineet and Burda, Yuri and Edwards, Harrison and Baker, Bowen and Lee, Teddy and Leike, Jan and Schulman, John and Sutskever, Ilya and Cobbe, Karl},
  booktitle={The Twelfth International Conference on Learning Representations},
  year={2023}
}

@article{wang2022self,
  title={Self-consistency improves chain of thought reasoning in language models},
  author={Wang, Xuezhi and Wei, Jason and Schuurmans, Dale and Le, Quoc and Chi, Ed and Narang, Sharan and Chowdhery, Aakanksha and Zhou, Denny},
  journal={arXiv preprint arXiv:2203.11171},
  year={2022}
}

@article{zelikman2022star,
  title={Star: Bootstrapping reasoning with reasoning},
  author={Zelikman, Eric and Wu, Yuhuai and Mu, Jesse and Goodman, Noah},
  journal={Advances in Neural Information Processing Systems},
  volume={35},
  pages={15476--15488},
  year={2022}
}

@inproceedings{besta2024graph,
  title={Graph of thoughts: Solving elaborate problems with large language models},
  author={Besta, Maciej and Blach, Nils and Kubicek, Ales and Gerstenberger, Robert and Podstawski, Michal and Gianinazzi, Lukas and Gajda, Joanna and Lehmann, Tomasz and Niewiadomski, Hubert and Nyczyk, Piotr and others},
  booktitle={Proceedings of the AAAI conference on artificial intelligence},
  volume={38},
  number={16},
  pages={17682--17690},
  year={2024}
}

@article{zelikman2024quiet,
  title={Quiet-star: Language models can teach themselves to think before speaking},
  author={Zelikman, Eric and Harik, Georges and Shao, Yijia and Jayasiri, Varuna and Haber, Nick and Goodman, Noah D},
  journal={arXiv preprint arXiv:2403.09629},
  year={2024}
}

@inproceedings{hao2023reasoning,
  title={Reasoning with language model is planning with world model},
  author={Hao, Shibo and Gu, Yi and Ma, Haodi and Hong, Joshua and Wang, Zhen and Wang, Daisy and Hu, Zhiting},
  booktitle={Proceedings of the 2023 Conference on Empirical Methods in Natural Language Processing},
  pages={8154--8173},
  year={2023}
}

@inproceedings{hsieh2023distilling,
  title={Distilling step-by-step! outperforming larger language models with less training data and smaller model sizes},
  author={Hsieh, Cheng-Yu and Li, Chun-Liang and Yeh, Chih-Kuan and Nakhost, Hootan and Fujii, Yasuhisa and Ratner, Alex and Krishna, Ranjay and Lee, Chen-Yu and Pfister, Tomas},
  booktitle={Findings of the Association for Computational Linguistics: ACL 2023},
  pages={8003--8017},
  year={2023}
}

@article{uesato2022solving,
  title={Solving math word problems with process-and outcome-based feedback},
  author={Uesato, Jonathan and Kushman, Nate and Kumar, Ramana and Song, Francis and Siegel, Noah and Wang, Lisa and Creswell, Antonia and Irving, Geoffrey and Higgins, Irina},
  journal={arXiv preprint arXiv:2211.14275},
  year={2022}
}

@article{turpin2023language,
  title={Language models don't always say what they think: Unfaithful explanations in chain-of-thought prompting},
  author={Turpin, Miles and Michael, Julian and Perez, Ethan and Bowman, Samuel},
  journal={Advances in Neural Information Processing Systems},
  volume={36},
  pages={74952--74965},
  year={2023}
}

@inproceedings{chen2021finqa,
  title={Finqa: A dataset of numerical reasoning over financial data},
  author={Chen, Zhiyu and Chen, Wenhu and Smiley, Charese and Shah, Sameena and Borova, Iana and Langdon, Dylan and Moussa, Reema and Beane, Matt and Huang, Ting-Hao and Routledge, Bryan R and others},
  booktitle={Proceedings of the 2021 Conference on Empirical Methods in Natural Language Processing},
  pages={3697--3711},
  year={2021}
}

@article{chen2022convfinqa,
  title={Convfinqa: Exploring the chain of numerical reasoning in conversational finance question answering},
  author={Chen, Zhiyu and Li, Shiyang and Smiley, Charese and Ma, Zhiqiang and Shah, Sameena and Wang, William Yang},
  journal={arXiv preprint arXiv:2210.03849},
  year={2022}
}

@article{gunasekar2023textbooks,
  title={Textbooks are all you need},
  author={Gunasekar, Suriya and Zhang, Yi and Aneja, Jyoti and Mendes, Caio C{\'e}sar Teodoro and Del Giorno, Allie and Gopi, Sivakanth and Javaheripi, Mojan and Kauffmann, Piero and de Rosa, Gustavo and Saarikivi, Olli and others},
  journal={arXiv preprint arXiv:2306.11644},
  year={2023}
}

@article{zhao2022multihiertt,
  title={MultiHiertt: Numerical reasoning over multi hierarchical tabular and textual data},
  author={Zhao, Yilun and Li, Yunxiang and Li, Chenying and Zhang, Rui},
  journal={arXiv preprint arXiv:2206.01347},
  year={2022}
}

@inproceedings{krumdick2024bizbench,
  title={Bizbench: A quantitative reasoning benchmark for business and finance},
  author={Krumdick, Michael and Koncel-Kedziorski, Rik and Lai, Viet Dac and Reddy, Varshini and Lovering, Charles and Tanner, Chris},
  booktitle={Proceedings of the 62nd Annual Meeting of the Association for Computational Linguistics (Volume 1: Long Papers)},
  pages={8309--8332},
  year={2024}
}

@article{xie2024finben,
  title={Finben: A holistic financial benchmark for large language models},
  author={Xie, Qianqian and Han, Weiguang and Chen, Zhengyu and Xiang, Ruoyu and Zhang, Xiao and He, Yueru and Xiao, Mengxi and Li, Dong and Dai, Yongfu and Feng, Duanyu and others},
  journal={Advances in Neural Information Processing Systems},
  volume={37},
  pages={95716--95743},
  year={2024}
}

@article{Chen2025StockBench,
	author = {Chen, Yanxu and Yao, Zijun and Liu, Yantao and Ye, Jin and Yu, Jianing and Hou, Lei and Li, Juanzi},
	doi = {10.48550/ARXIV.2510.02209},
	year = {2025},
	publisher = {arXiv},
	title = {StockBench: Can {LLM} {Agents} {Trade} {Stocks} {Profitably} {In} {Real}-world {Markets}?},
	url = {https://arxiv.org/abs/2510.02209},
}

@inproceedings{li2024alphafin,
  title={Alphafin: Benchmarking financial analysis with retrieval-augmented stock-chain framework},
  author={Li, Xiang and Li, Zhenyu and Shi, Chen and Xu, Yong and Du, Qing and Tan, Mingkui and Huang, Jun},
  booktitle={Proceedings of the 2024 joint international conference on computational linguistics, language resources and evaluation (LREC-COLING 2024)},
  pages={773--783},
  year={2024}
}

@article{hurst2024gpt,
  title={Gpt-4o system card},
  author={Hurst, Aaron and Lerer, Adam and Goucher, Adam P and Perelman, Adam and Ramesh, Aditya and Clark, Aidan and Ostrow, AJ and Welihinda, Akila and Hayes, Alan and Radford, Alec and others},
  journal={arXiv preprint arXiv:2410.21276},
  year={2024}
}

@article{comanici2025gemini,
  title={Gemini 2.5: Pushing the frontier with advanced reasoning, multimodality, long context, and next generation agentic capabilities},
  author={Comanici, Gheorghe and Bieber, Eric and Schaekermann, Mike and Pasupat, Ice and Sachdeva, Noveen and Dhillon, Inderjit and Blistein, Marcel and Ram, Ori and Zhang, Dan and Rosen, Evan and others},
  journal={arXiv preprint arXiv:2507.06261},
  year={2025}
}

@misc{anthropic2025claude37,
  title       = {Claude 3.7 Sonnet System Card},
  author      = {{Anthropic}},
  year        = {2025},
  month       = {February},
  url         = {https://www.anthropic.com/claude-3-7-sonnet-system-card},
  note        = {Released February 23, 2025}
}

@misc{meta2025llama,
  title       = {Llama 4 Scout Model Card},
  author      = {{Meta}},
  year        = {2025},
  month       = {April},
  url         = {https://github.com/meta-llama/llama-models/blob/main/models/llama4/MODEL_CARD.md},
}

@article{dettmers2023qlora,
  title={Qlora: Efficient finetuning of quantized llms},
  author={Dettmers, Tim and Pagnoni, Artidoro and Holtzman, Ari and Zettlemoyer, Luke},
  journal={Advances in neural information processing systems},
  volume={36},
  pages={10088--10115},
  year={2023}
}

@misc{Meta_2024b, title={Meta Llama 3.1 Model Information}, url={https://github.com/meta-llama/llama-models/blob/main/models/llama3_1/MODEL_CARD.md}, author={Meta}, year={2024}, month={Jul}}

@misc{OpenAI_2025, url={https://openai.com/index/gpt-5-1/}, journal={GPT-5.1: A smarter, more conversational chatgpt | openai}, author={{OpenAI et al.}}, year={2025}, month={Nov}}

@misc{Claude_2025, url={https://assets.anthropic.com/m/12f214efcc2f457a/original/Claude-Sonnet-4-5-System-Card.pdf}, journal={System Card: Claude Sonnet 4.5}, author={{Anthropic et al.}}, year={2025}, month={Sep}}

@article{DeepSeek2024DeepSeek,
	author = {{DeepSeek-AI} and Liu, Aixin and Feng, Bei and Xue, Bing and Wang, Bingxuan and Wu, Bochao and Lu, Chengda and Zhao, Chenggang and Deng, Chengqi and Zhang, Chenyu and Ruan, Chong and Dai, Damai and Guo, Daya and Yang, Dejian and Chen, Deli and Ji, Dongjie and Li, Erhang and Lin, Fangyun and Dai, Fucong and Luo, Fuli and Hao, Guangbo and Chen, Guanting and Li, Guowei and Zhang, H. and Bao, Han and Xu, Hanwei and Wang, Haocheng and Zhang, Haowei and Ding, Honghui and Xin, Huajian and Gao, Huazuo and Li, Hui and Qu, Hui and Cai, J. L. and Liang, Jian and Guo, Jianzhong and Ni, Jiaqi and Li, Jiashi and Wang, Jiawei and Chen, Jin and Chen, Jingchang and Yuan, Jingyang and Qiu, Junjie and Li, Junlong and Song, Junxiao and Dong, Kai and Hu, Kai and Gao, Kaige and Guan, Kang and Huang, Kexin and Yu, Kuai and Wang, Lean and Zhang, Lecong and Xu, Lei and Xia, Leyi and Zhao, Liang and Wang, Litong and Zhang, Liyue and Li, Meng and Wang, Miaojun and Zhang, Mingchuan and Zhang, Minghua and Tang, Minghui and Li, Mingming and Tian, Ning and Huang, Panpan and Wang, Peiyi and Zhang, Peng and Wang, Qiancheng and Zhu, Qihao and Chen, Qinyu and Du, Qiushi and Chen, R. J. and Jin, R. L. and Ge, Ruiqi and Zhang, Ruisong and Pan, Ruizhe and Wang, Runji and Xu, Runxin and Zhang, Ruoyu and Chen, Ruyi and Li, S. S. and Lu, Shanghao and Zhou, Shangyan and Chen, Shanhuang and Wu, Shaoqing and Ye, Shengfeng and Ye, Shengfeng and Ma, Shirong and Wang, Shiyu and Zhou, Shuang and Yu, Shuiping and Zhou, Shunfeng and Pan, Shuting and Wang, T. and Yun, Tao and Pei, Tian and Sun, Tianyu and Xiao, W. L. and Zeng, Wangding and Zhao, Wanjia and An, Wei and Liu, Wen and Liang, Wenfeng and Gao, Wenjun and Yu, Wenqin and Zhang, Wentao and Li, X. Q. and Jin, Xiangyue and Wang, Xianzu and Bi, Xiao and Liu, Xiaodong and Wang, Xiaohan and Shen, Xiaojin and Chen, Xiaokang and Zhang, Xiaokang and Chen, Xiaosha and Nie, Xiaotao and Sun, Xiaowen and Wang, Xiaoxiang and Cheng, Xin and Liu, Xin and Xie, Xin and Liu, Xingchao and Yu, Xingkai and Song, Xinnan and Shan, Xinxia and Zhou, Xinyi and Yang, Xinyu and Li, Xinyuan and Su, Xuecheng and Lin, Xuheng and Li, Y. K. and Wang, Y. Q. and Wei, Y. X. and Zhu, Y. X. and Zhang, Yang and Xu, Yanhong and Xu, Yanhong and Huang, Yanping and Li, Yao and Zhao, Yao and Sun, Yaofeng and Li, Yaohui and Wang, Yaohui and Yu, Yi and Zheng, Yi and Zhang, Yichao and Shi, Yifan and Xiong, Yiliang and He, Ying and Tang, Ying and Piao, Yishi and Wang, Yisong and Tan, Yixuan and Ma, Yiyang and Liu, Yiyuan and Guo, Yongqiang and Wu, Yu and Ou, Yuan and Zhu, Yuchen and Wang, Yuduan and Gong, Yue and Zou, Yuheng and He, Yujia and Zha, Yukun and Xiong, Yunfan and Ma, Yunxian and Yan, Yuting and Luo, Yuxiang and You, Yuxiang and Liu, Yuxuan and Zhou, Yuyang and Wu, Z. F. and Ren, Z. Z. and Ren, Zehui and Sha, Zhangli and Fu, Zhe and Xu, Zhean and Huang, Zhen and Zhang, Zhen and Xie, Zhenda and Zhang, Zhengyan and Hao, Zhewen and Gou, Zhibin and Ma, Zhicheng and Yan, Zhigang and Shao, Zhihong and Xu, Zhipeng and Wu, Zhiyu and Zhang, Zhongyu and Li, Zhuoshu and Gu, Zihui and Zhu, Zijia and Liu, Zijun and Li, Zilin and Xie, Ziwei and Song, Ziyang and Gao, Ziyi and Pan, Zizheng},
	doi = {10.48550/ARXIV.2412.19437},
	year = {2024},
	publisher = {arXiv},
	title = {DeepSeek-{V3} {Technical} {Report}},
	url = {https://arxiv.org/abs/2412.19437},
}

@article{yang2025qwen3,
  title={Qwen3 technical report},
  author={Yang, An and Li, Anfeng and Yang, Baosong and Zhang, Beichen and Hui, Binyuan and Zheng, Bo and Yu, Bowen and Gao, Chang and Huang, Chengen and Lv, Chenxu and others},
  journal={arXiv preprint arXiv:2505.09388},
  year={2025}
}

@article{Jiang2023Mistral,
	author = {Jiang, Albert Q. and Sablayrolles, Alexandre and Mensch, Arthur and Bamford, Chris and Chaplot, Devendra Singh and Casas, Diego de las and Bressand, Florian and Lengyel, Gianna and Lample, Guillaume and Saulnier, Lucile and Lavaud, L{\' e}lio Renard and Lachaux, Marie-Anne and Stock, Pierre and Scao, Teven Le and Lavril, Thibaut and Wang, Thomas and Lacroix, Timoth{\' e}e and Sayed, William El},
	doi = {10.48550/ARXIV.2310.06825},
	year = {2023},
	publisher = {arXiv},
	title = {Mistral 7B},
	url = {https://arxiv.org/abs/2310.06825},
}

@INPROCEEDINGS{10993716,
  author={Wang, Liujianfu and Du, Yuyang and Lin, Jingqi and Chen, Kexin and Liew, Soung Chang},
  booktitle={2025 International Conference on Computing, Networking and Communications (ICNC)}, 
  title={Rephrase and Contrast: Fine-Tuning Language Models for Enhanced Understanding of Communication and Computer Networks}, 
  year={2025},
  volume={},
  number={},
  pages={588-594},
  doi={10.1109/ICNC64010.2025.10993716}}

\end{document}